\documentclass[runningheads]{template/llncs}

\usepackage{mwe}
\usepackage{lipsum}

\usepackage{float}
\usepackage{physics}
\usepackage{booktabs}
\usepackage{multirow}
\usepackage{hyperref}
\usepackage{tablefootnote}
\usepackage{subcaption}

\usepackage{comment}
\usepackage{pifont}
\usepackage{amsmath}
\usepackage{amsfonts}
\usepackage{amssymb}
\usepackage{mathtools}
\usepackage{xcolor}
\usepackage{textcomp}

\usepackage{subcaption}
\usepackage{colortbl}
\usepackage{graphicx}

\usepackage{adjustbox}
\usepackage{blindtext}
\usepackage{tikz}

\newcommand*{\mybox}[1]{\framebox{\strut #1}}

\begin{document}
	\title{Few-Sample Named Entity Recognition for Security Vulnerability Reports by Fine-Tuning Pre-Trained Language Models}
	\titlerunning{Few-Sample NER for Security Vulnerability Reports by Fine-Tuning PLMs}
	
	\author{
			Guanqun Yang\inst{1}\and 
			Shay Dineen\inst{1}\and 
			Zhipeng Lin\inst{1} \and 
			Xueqing Liu\inst{1}
		 	}
	\authorrunning{G. Yang et al.}
	\institute{
		Department of Computer Science,\\ Stevens Institute of Technology, Hoboken NJ 07030, USA\\
		\email{\{gyang16, sdineen1, zlin30, xliu127\}@stevens.edu}
	}
	\maketitle
	\begin{abstract}
	Public security vulnerability reports (e.g., CVE reports) play an important role in the maintenance of computer and network systems. 
	Security companies and administrators rely on information from these reports to prioritize tasks on developing and deploying patches to their customers. 
	Since these reports are unstructured texts, automatic information extraction (IE) can help scale up the processing by converting the unstructured reports to structured forms, e.g., software names and versions~\cite{Dong2019TowardsTD} and vulnerability types~\cite{sun2021generating}. 
	Existing works on automated IE for security vulnerability reports often rely on a large number of labeled training samples ~\cite{Dong2019TowardsTD,gasmi2019information,Zhou2020ImprovingSB}. 
	However, creating massive labeled training set is both expensive and time consuming. 
	In this work, for the first time, we propose to investigate this problem where only a small number of labeled training samples are available. 
	In particular, we investigate the performance of fine-tuning several state-of-the-art pre-trained language models on our small training dataset. 
	The results show that with pre-trained language models and carefully tuned hyperparameters, we have reached or slightly outperformed the state-of-the-art system~\cite{Dong2019TowardsTD} on this task. 
    Consistent with previous two-step process of first fine-tuning on main category and then transfer learning to others as in \cite{Dodge2020FineTuningPL}, if otherwise following our proposed approach, the number of required labeled samples substantially decrease in both stages: 90\% reduction in fine-tuning from 5758 to 576, and 88.8\% reduction in transfer learning with 64 labeled samples per category.   
	Our experiments thus demonstrate the effectiveness of few-sample learning on NER for security vulnerability report.
	This result opens up multiple research opportunities for few-sample learning for security vulnerability reports, which is discussed in the paper. 
	Our implementation for few-sample vulnerability entity tagger in security reports could be found at \href{https://github.com/guanqun-yang/FewVulnerability}{https://github.com/guanqun-yang/FewVulnerability}.
	
	\end{abstract}

	\section{Introduction}

Security vulnerabilities pose great challenges to software and network systems. 
For example, there are numerous reported data breaches of Uber, Equifax, Marriott, and Facebook that jeopardize hundreds of millions of customers' information security \cite{gaglione2019equifax};
vulnerable in-flight automation software used in Boeing 737-MAX are found to be guilty for a series of crashes in 2018 \cite{Palmer2020}. 
To allow the tracking and management of security vulnerabilities, people have created public security vulnerability reports and store them in vulnerability databases for reference. 
For example, the Common Vulnerabilities and Exposures (CVE) database is created and maintained by MITRE \footnote{\href{https://www.mitre.org/}{\texttt{https://www.mitre.org/}}}; the National Vulnerability Database (NVD) is another database maintained by the U.S. government. 
These databases have largely facilitated security companies and vendors to manage and prioritize deployment of patches. 

To scale up the management of vulnerability entries, existing works have leveraged natural language processing (NLP) techniques to extract information from unstructured vulnerability reports. 
For example, software name, software version, and vulnerability type.
The extraction of such information can help accelerate security administration in the following scenarios: First, after converting the unstructured reports into structured entries such as software name and software version, such information can be directly used for detecting the inconsistencies between different records of vulnerability reports. 
For example, significant inconsistency was detected between reports from the CVE and NVD databases~\cite{Dong2019TowardsTD}, which may cause system administrators to retrieve outdated or incorrect security alerts, exposing systems under their watch to hazard; inconsistency was also detected between two entries created by the same vendor Samsung in~\cite{farhang2020empirical}; Second, the extracted entries can be used in other downstream applications. 
For example, we can leverage different categories of entries (e.g., vendor name, vulnerability types, attacker name) to construct a knowledge ontology for modeling the interplay between security entries~\cite{gao2021system,bridges2013automatic}. 
The extracted entries have also been used for automatically generating a natural language summarization of the original report by leveraging a template~\cite{sun2021generating}. 

The scalable processing of vulnerability reports into structured forms relies on the support from an effective information extraction (IE) system. 
With the massive amount of reports, manual extraction of information is intractable and most of the existing systems rely on machine learning approaches. 
For example, Mulwad et al.~\cite{mulwad2011extracting} built an SVM-based system which extracted security concepts from web texts. 
In the recent years, with the rise of deep neural networks, systems based on deep learning were proposed, e.g., by using LSTM~\cite{gasmi2019information,Zhou2020ImprovingSB} and bi-directional GRU~\cite{Dong2019TowardsTD}. 
Deep learning has been shown to be effective in this task, e.g., with an F1 score of 98\%~\cite{Dong2019TowardsTD}. 


Nevertheless, a major challenge of applying existing systems (e.g., recurrent neural network like GRU ~\cite{Dong2019TowardsTD} and LSTM ~\cite{gasmi2019information,Zhou2020ImprovingSB}) is that they often require a large number of labeled training samples to perform well;
this number could range from a few thousand \cite{Zhang2020} to tens of thousand \cite{gasmi2019information}.
Labeling a NER dataset specific to computer security domain at this scale is both costly and time-consuming.
This problem of annotation is further exacerbated by the fact that new security reports are routinely generated.
Therefore, it is vital to investigate the possibility to alleviate the burden of labeling huge corpus, i.e., few-sample learning.

In this paper, for the first time, we investigate the problem of information extraction for public vulnerability reports under the setting where only a small number of labeled training samples are available. 
Our goal is to minimize the number of labeled samples without suffering from a significant performance degradation.

In recent years, significant progress has been made for few-shot learning, e.g., image classification~\cite{finn2017model}, text classification~\cite{Gao2020}, text generation~\cite{brown2020language}, etc. 
In these tasks, the training set and validation set often only contain ten or fewer examples, and it has been shown that the available small number of examples are all what the model needs to achieve a good performance~\cite{chang2018pyramid}. 
To generalize to unseen examples, few-shot learning approaches often leverage meta-learning~\cite{finn2017model}, metric learning~\cite{snell2017prototypical,chen2021non} and model pre-training. 
For example, the GPT-3 model~\cite{brown2020language} contains 175 billion parameters and was pre-trained on 45TB of unlabeled text data. 
It is able to accomplish a variety of generative tasks (e.g., question answering, conversation generation, text to command) when the user only provides a few examples for demonstration. 
For information extraction, more specifically, named entity recognition (NER), several existing works study its few-shot setting, where only tens of labeled samples are provided during training ~\cite{Yang2020SimpleAE,hou2020few}. 
However, NER systems to date can only achieve a 62\% F1 score (evaluated on the CoNLL 2003 dataset~\cite{sang2003conll}). 
When applied to the security vulnerability reports domain, it is thus unclear whether the performance can be satisfactory enough to security administrators and vendors, as a lower error rate is often required~\cite{Dong2019TowardsTD}. 
As a result, in this paper, we investigate the following research questions:

\begin{itemizebullet}
    \item Is it possible to match the state-of-the-art performance in NER for security vulnerability reports by using only a small number of labeled examples?
    \item What is the minimum number of labeled examples required for this problem?
\end{itemizebullet}

To answer these research questions, we conduct an experimental study on \emph{few-sample} named entity recognition for vulnerability reports\footnote{Here we frame our problem under the term "few-sample" learning instead of "few-shot" learning because our approach generally requires tens to a few hundred labeled training samples, which is more than "few-shot". We adopt the name "few-sample" from \cite{Zhang2020}. }, which to the best of our knowledge has not been explored by previous works.
As an initial study, we focus on the \texttt{VIEM} dataset~\cite{Dong2019TowardsTD}, a dataset of vulnerability reports containing 3 types of entities (i.e., \emph{software name, software version, outside}) and 13 sub-datasets based on the category of vulnerabilities. 
For the largest sub-dataset, we find that through fine-tuning pre-trained language models, it is possible to match the state-of-the-art performance (i.e., 0.92 to 0.93 F1 score for different entity types~\cite{Dong2019TowardsTD}) using only 576 labeled sentences for the training dataset, or 10\% of the original training data. 
For the other 12 categories, the same approach matches the SOTA performance with only 11.2\% of the labeled samples. 
Our experiments also show that the simple fine-tuning approach works better than state-of-the-art systems for few-shot NER~\cite{Yang2020SimpleAE}. 

The main contributions of this paper are:

\begin{itemizebullet}
    \item We propose, for the first time, to study the problem of named entity recognition (NER) for security vulnerability reports;
    \item We perform an experimental study by fine-tuning three state-of-the-art language models and evaluating one metric learning-based method for few-sample NER in vulnerability reports, which shows that the fine-tuning approach can achieve the similar performance using only 10\% or 11.2\% of the original training dataset;
    \item We discuss multiple future research opportunities of information extraction for security vulnerability reports;
\end{itemizebullet}

The rest of this paper is organized as follows. Section~\ref{sec:report} defines the problem and discusses the domain and data-specific challenges; 
Section~\ref{sec:method} introduces the methods examined; Section~\ref{sec:exp} describes the experimental steps, datasets, and experimental results; 
Section~\ref{sec:relwork} analyzes the related work; finally, Section~\ref{sec:conclusion} draws the conclusion, analyzes the limitation of this work and discusses several future directions of research for few-sample learning for security vulnerability reports. 

	\section{Problem Definition and Challenges}\label{sec:report}

\subsection{Few-Sample Named Entity Recognition}
\label{sec:setting}
Name entity recognition (NER) is the task of assigning predefined entity names to tokens in the input text.
Formally, with predefined tagging set $\mathcal{Y}$ and a sequence of tokens (input sentence) $X_i=[x_{i1}, x_{i2}, \cdots, x_{iT}]$, each $x_{ik}\ (k=1,2,\cdots, T)$ corresponds to a tag $y_{ik} \in \mathcal{Y}$ specifying the its entity type. 
This gives $X_i$ a tagging sequence $Y_i=[
y_{i1}, y_{i2}, \cdots, y_{iT}]$ and a typical dataset of NER is of the form $\mathcal{D}=\{(X_i, Y_i)\}_{i=1}^N$.

Notice there exist two distinct definitions for what a few-shot learning algorithm tries to achieve~\cite{Wang2019GeneralizingFA}. In the first definition, the goal is to match the performance when only a subset of training data is available; in the second definition, the test set contains unseen classes in the training set and the algorithm needs to generalize to the unseen classes. In this work, we focus on the first setting. We refer readers interested in the details of these two settings to \cite{Zhang2020}.

\subsection{Named Entity Recognition for Vulnerability Reports}
\begin{figure}[t]
    \centering
    \includegraphics[width=\textwidth]{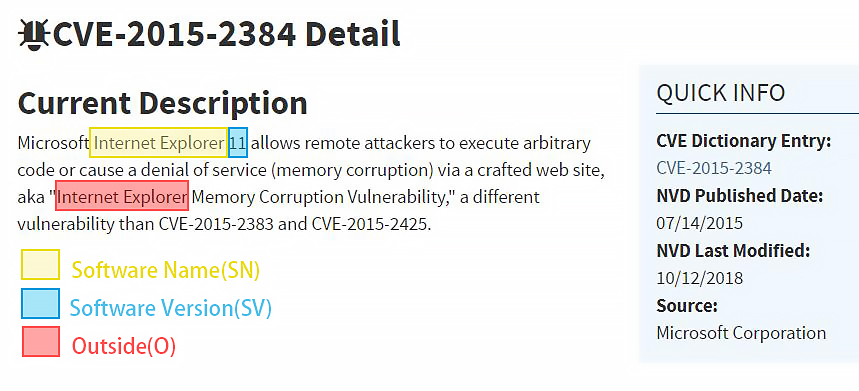}
    \caption{An example of public security vulnerability report }
    \label{fig:example}
\end{figure}

An example of security vulnerability report (CVE-2015-2384) is shown in Figure~\ref{fig:example}, which is a snapshot from the NVD website\footnote{\href{https://nvd.nist.gov}{\texttt{https://nvd.nist.gov}}}. 
In this paper, we focus on the \texttt{VIEM} dataset from \cite{Dong2019TowardsTD}, which contains 13 categories of vulnerabilities and two types of named entities: software name and software version. 
That is, the tag set $\mathcal{Y}$ is instantiated as $\{\texttt{SN}, \texttt{SV}, \texttt{O}\}$, where \texttt{SN}, \texttt{SV}, \texttt{O} refer to software name, software version, and non-entities, respectively.
For example, as is shown in Figure~\ref{fig:example}, the report CVE-2015-2384 warrants a security alert where the software \texttt{Internet Explorer} of version \texttt{11} might expose its vulnerability to attackers.
The training, validation, and testing set share the same tag set $\mathcal{Y}$.

\subsection{Data-Specific Challenges}\label{sec-challenge}
There exists several data-specific challenges in the \texttt{VIEM} dataset, which we summarize as below. 

\subsubsection{Contextual Dependency} 

When identifying vulnerable software names and versions in vulnerability reports, the tagger has to take the entire context into account. 
For example, as is shown in Figure \ref{fig:example} (CVE-2015-2384), \texttt{Internet Explorer} is tagged as a software name for its first occurrence; in the second occurrence, it appears in the vulnerability name \emph{Internet Explorer Memory Corruption Vulnerabilities} thus is no longer a software name.
Software version tag is also context dependent: in the examples below, \texttt{1.0} is tagged \texttt{SV} in CVE-2002-2323 and CVE-2010-3487 but \texttt{O} in CVE-2006-4710, this is because when the software name (James M. Snell) is tagged \texttt{O}, the software version should also be tagged \texttt{O}.

\begin{itemizebullet}
    \item CVE-2002-2323: \textit{\colorbox{yellow!30}{Sun PC NetLink} \colorbox{cyan!30}{1.0} through 1.2 \ldots  access restriction.}
    \item CVE-2010-3487: \textit{Directory traversal vulnerability in \colorbox{yellow!30}{YelloSoft Pinky} \colorbox{cyan!30}{1.0} \ldots in the URL.}
    \item CVE-2006-4710: \textit{Multiple cross-site scripting \ldots as demonstrated by certain test cases of the \colorbox{red!30}{James M. Snell Atom} \colorbox{red!30}{1.0} feed reader test suite.}
\end{itemizebullet}


\subsubsection{Data Imbalance}
The goal of NER system is to identify entities of interest and tag all the other tokens as outsider. 
It is therefore possible that the entire sentence includes only outsider tokens.
As is shown in Table \ref{tab:imbalance}, in the general domain NER dataset like CoNLL 2003 and OntoNotes 5, the proportion of such sentences in training set is relatively low. 
On the other hand, in the vulnerability reports, the proportion of all-\texttt{O} sentences is high.
Such data imbalance problem can cause the classifier to trade off minority group's performance for that of majority group, leading to a higher error rate for minority samples~\cite{Johnson2019SurveyOD}.

\begin{table}
	\centering
	\caption[imbalance]{The proportion of sentences in training set that include only non-entities. The statistics of CoNLL 2003 and OntoNotes 5 are estimated from publicly available dataset releases. In the \texttt{VIEM} dataset, the vulnerability category \texttt{memc} is intentionally balanced in \cite{Dong2019TowardsTD} to ease transfer learning, making the aggregate (i.e. \texttt{VIEM} (all)) non-entity-only proportion lower.}
	\begin{tabular}{ccc}
	\hline
	   Dataset& Domain& Non-entity-only sentences\\
	\hline
	CoNLL 2003\tablefootnote{\url{https://huggingface.co/datasets/conll2003}}
	& General& 20.71\%\\
   OntoNotes 5\tablefootnote{\url{https://github.com/Fritz449/ProtoNER/tree/master/ontonotes}}& General& 51.60\%\\
	\texttt{VIEM} (all)& Security& 60.34\%\\
	\texttt{VIEM} (w/o \texttt{memc})&Security& 75.44\%\\
	\hline
	\end{tabular}
	\label{tab:imbalance}
\end{table}
	\section{Few-Sample NER for Vulnerability Reports}
\label{sec:method}

In this section, we introduce the methods investigated in this paper. 
The first method leverages fine-tuning  pre-trained language models (PLM); the second method is a few-shot NER method based on the nearest neighbor approach~\cite{Yang2020SimpleAE}. 

\subsection{Fine-Tuning Pre-trained Language Models with Hundreds of Training Labels}

\subsubsection{Fine-Tuning Framework and Selection of Pre-trained Language Models}\label{sec:choose-model}
Before we describe our approach, first we introduce our choices for the language model to be fine-tuned. There exists hundreds of candidate models from the HugginFace model hub\footnote{\href{https://huggingface.co/models}{\texttt{https://huggingface.co/models}}}. 
We investigate three models out of all: \texttt{bert-large-cased}, \texttt{roberta-large}, and \texttt{electra-base-\allowbreak discriminator}.
BERT was the first pre-trained language model that achieved great performance compared to prior art; RoBERTa and Electra are among the top-ranked language models on the GLUE leaderboard as of Jan 2021 \footnote{\href{https://gluebenchmark.com/}{\texttt{https://gluebenchmark.com/}}}. 
For BERT, we choose the cased version \texttt{bert-large-cased} rather than the uncased counterpart (i.e. \texttt{bert-large-uncased}) as token casings could be informative for the identification of software names (\texttt{SN}).
Our implementation is based on the HugginFace \texttt{transformers} library (4.5.1), more specifically, we leverage the \texttt{AutoModel\allowbreak ForTokenClassification} class for fine-tuning the 13 categories in the \texttt{VIEM} dataset. 

\subsubsection{Fine-Tuning on the \texttt{memc} Category} 
Following the experimental protocol of \texttt{VIEM} system ~\cite{Dong2019TowardsTD}, the PLM in our system is first fine-tuned on the vulnerability category of \texttt{memc} (short for "memory corruption") as, from the statistics of the dataset shown in Table \ref{tab:stat}, this category suffers less from data imbalance described in section \ref{sec:report}; fine-tuning on this category could therefore best help PLMs learn representations specific to computer security domain.
However, rather than allowing PLMs accessing the entire training dataset of \texttt{memc}, we constrain its access to the training data from 1\% to 10\% of the entire training set.

\subsubsection{Transfer Learning to the Other 12 Categories}
Previous works on fine-tuning language models show that fine-tuning can benefit from multi-task learning or transfer learning, i.e., fine-tuning on a related task before fine-tuning on the objective task~\cite{phang2018sentence}.
We adopt this approach, i.e., the PLM is first fine-tuned on the \texttt{memc} category, then we start from the checkpoint and continue the fine-tuning on the 12 categories. 
We again limit the number of annotated samples during transfer learning.
Throughout the transfer learning experiment, we sample varying number of training samples from each of the 12 categories, and we combine the 12 subsets to form an \emph{aggregate} training set.
For validation and testing, however, we evaluate the model with transfer learning on each category \emph{independently}.

\subsubsection{Hyperparameter Optimization}
The hyperparameters for fine-tuning includes the learning rate, batch size, and the number of training epochs.
Inappropriately setting one of these hyperparameter might cause performance degradation~\cite{Mosbach2020}. 
As a result, hyperparameter tuning is critical as they could have substantial impact on the model performance.

\subsubsection{Setting the Random Seed}\label{sec:ft-random}
The fine-tuning performance for PLM are known to be unstable, especially under the typical few-sample setting.
Some seemingly benign parameters, like choice of random seed, could introduce sizable variance on the model performance.
Indeed, the seed choice could influence weight initialization for task-specific layers when applying PLMs for downstream task~\cite{Dodge2020FineTuningPL}.
When these task-specific layers are initialized poorly, coupled with small size of the training set, it is unlikely for the network to converge to a good solution without extensive training.
In fact, previous work has shown that some seeds are consistently better than others for network optimization~\cite{Dodge2020FineTuningPL}.
But it is not possible to know a priori which seed is better than the others.

\subsection{Few-shot Named Entity Recognition}\label{sec:structshot}
After trying the fine-tuning methods above, we find that these methods still require hundreds of training examples, one question is whether it is possible to further reduce the number of training labels, so that only a few training labels are required. 
For example, the system designed by Yang and Katiyar \cite{Yang2020SimpleAE}, named \texttt{StructShot}, manages to beat competitive systems with a few samples, i.e. only 1 to 5 annotated named entities are required for each named entity class.

The key technique used in \texttt{StructShot} is the nearest neighbor search.
Specifically, rather than directly training a PLM with token classification objective as we do,
their system is first fine-tuned on a high-resource yet distantly related dataset; then they use the model as a feature extractor to obtain token representations of each of tokens in the test set and a small support set, both coming from low-resource task under investigation.
With these two sets of representations in hand, test token's tag is determined through retrieving the most similar token's tag in the support set.
A CRF decoding procedure follows the similarities estimated from previous steps to give the optimal tag sequence.



	\section{Experiments}
\label{sec:exp}

\subsection{Datasets}\label{sec:dataset}
We experiment on the \texttt{VIEM} dataset provided by \cite{Dong2019TowardsTD}. 
This dataset is collected from the public vulnerability databases Common Vulnerabilities and Exposure (CVE) and numerous security forums.
The goal of this dataset is to identify vulnerable software names and software versions.
Therefore, the tag set $\mathcal{Y}$ includes \texttt{SN} (software names) and \texttt{SV} (software versions).
Similar to the other NER datasets, the \texttt{O} (outside) tag is used to represent all other tokens except vulnerable software names and versions.
The dataset is manually labeled by three security experts in hope of minimizing labeling errors and enforcing consistent annotation standards.
The records in dataset range in 20 years and contain all 13 categories listed on CVE website.
The statistics of the dataset is shown in Table \ref{tab:stat}.
We only report the average statistics of 12 categories here, and we refer readers interested in detailed statistics of these categories to Table \ref{tab:stat-detail}. 
From the statistics of the dataset shown in Table \ref{tab:stat}, we observe the \emph{dataset imbalance}, a challenge discussed in section \ref{sec-challenge}, in different levels.
\begin{itemizebullet}
	\item \textbf{Sample Number Imbalance} The \texttt{memc} category has significantly more samples in every split than the other 12 categories. 
	\item \textbf{Entity Number Imbalance} The \texttt{memc} category has a predominately higher entity proportion in both sentence and token level than the other 12 categories.
	\item \textbf{Entity Type Imbalance} The appearances of \texttt{SV} tokens are generally more frequent than \texttt{SN}; sometimes this difference could vary up to 4.38\% (see valid split of \texttt{memc}). 
\end{itemizebullet}

\begin{table}[h]
    \centering
	\caption{Statistic of the \texttt{VIEM} dataset.}
	\begin{tabular}{lcccccc}
		\toprule
		Category & \multicolumn{3}{c}{\texttt{memc}} & \multicolumn{3}{c}{Average of other 12 categories} \\
		Split &      train &      valid &       test &     train &     valid &      test \\
		\midrule
		Number of sample                            &  5758 &  1159 &  1001 &  468.00 &  116.67 &  556.25 \\
		Sentence-level entity proportion            &     0.5639 &     0.3287 &     0.4555 &    0.2435 &    0.2677 &    0.2620 \\
		Token-level entity proportion (\texttt{SN}) &     0.0613 &     0.0368 &     0.0559 &    0.0214 &    0.0240 &    0.0237 \\
		Token-level entity proportion (\texttt{SV}) &     0.0819 &     0.0807 &     0.0787 &    0.0308 &    0.0331 &    0.0331 \\

	\bottomrule
	\end{tabular}
	\label{tab:stat}
\end{table}

\subsubsection{Preparing for the Few-Sample Dataset}
Among the 13 vulnerability categories in the dataset official release, only \texttt{memc} includes official validation split.
In order to enable hyperparameter optimization and model selection on the other 12 categories, we randomly sample 10\% of the training samples as the held-out validation set.

As we strive to investigate the performance of PLMs with the reduced number of annotated samples after fine-tuning and transfer learning, we create the training and validation set by sampling a subset of the full dataset.
More specifically, for training set:

\begin{itemizebullet}
    \item \textbf{Fine-Tuning} We vary the proportion of sampled data from the \texttt{memc} training set from 1\% to 10\% (equivalent to a sample size from 58 to 576) and investigate the the proportion to measure the adaptation of PLM to the computer security domain.
    \item \textbf{Transfer Learning} We vary the number of samples in the other 12 categories from $\{32, 64, 128, 256\}$ per category.
    Compared to fine-tuning, we retrain from sampling by proportion considering the fact that there are much smaller training set for the other 12 categories than the \texttt{memc} category.
\end{itemizebullet}

For the validation set, we make sure it has the same size as the training set whenever possible: the infeasible cases are those when the size of the sampled training set exceeds the 10\% limit.
This comes from the consideration that the validation samples also require annotations and therefore consume labeling budget.
Apart from the conversion of text form into data format required by model training, 
we do not perform any additional data processing to make sure the comparison is fair with previous system.
Throughout the data preparation, we fix the random seed to 42.

\subsection{Evaluation Metrics}
Following the evaluation metrics used in \cite{Dong2019TowardsTD}, we use the precision, recall, and F1 score to evaluate our NER system, the definitions of precision, recall and F1 score are standard.

\subsection{Experimental Setup}
\subsubsection{Hyperparameter Optimization Settings}
We leverage the built-in API for hyperparameter optimization from the HugginFace library~\cite{Wolf2019HuggingFacesTS}. 
We leverage the grid search where the search space is as follows: the learning rate is selected from $\{1e-6, 5e-6, 1e-5\}$; the number of training epochs is chosen from $\{3, 5\}$; the batch size, is fixed to 2 per device because we find the resulting additional training iterations favorable as compared to setting it to 4 or 8 in our pilot experiment. 
All the other hyperparameters values are fixed to the default values in HuggingFace's \texttt{transformers-4.5.1}. 
Besides the grid search, we also experiment with the Bayesian optimization, but it fails to outperform the grid search with same computational resources. 
We fix the seed for model training to 42. 
We randomly restart every experiment 10 times and report the average score. 
We use the half-precision (i.e. \texttt{fp16}) mode to accelerate the training. 
During the hyperparameter optimization, we select the best checkpoint within each trial. 
For trials with different epoch numbers, we partition the training iterations so that the same number of model checkpoints are saved during training across different experiments.
Due to the class imbalance problem (Section~\ref{sec:dataset}), measuring checkpoint quality solely based on metric of the \texttt{SN} or \texttt{SV} is suboptimal; therefore, we weigh their F1 score based on the number of their appearances in the groundtruth, i.e. $N_{\texttt{SN}}$ and $N_{\texttt{SV}}$.
\begin{equation*}
\bar{F}_1 = 
\frac{N_{\texttt{SN}}F_{\texttt{SN}}+N_{\texttt{SV}}F_{\texttt{SV}}}{N_{\texttt{SN}}+N_{\texttt{SV}}}
\end{equation*}
This single evaluation metric provides a tradeoff between metrics on different entities during model selection.

\subsubsection{Evaluated Methods}\label{sec:ss}
Our experiments contain two parts: first, we evaluate the performance of few-sample learning on the \texttt{memc} category; second, we evaluate the performance of few-sample learning on the other 12 categories. 

For the \texttt{memc} category, we fine-tune the models with varying proportions of the \texttt{memc} training data from 1\% to 10\%.
We are interested to know whether fine-tuning with a small proportion of samples is feasible for PLMs to reach the performance of competitive system like \texttt{VIEM}.

For the other 12 categories, we compare the following 4 settings:
\begin{itemizebullet}
    \item \textbf{FT} This setting is the direct application of models fine-tuned on \texttt{memc} to the other categories.
    We expect the performance of \texttt{FT} to be the lowest across 4 settings.
	\item \textbf{FT+SS} This setting is consistent with the way this system is used in \cite{Yang2020SimpleAE}:
	after the PLM is fine-tuned on the \texttt{memc} category, it is coupled with \texttt{StructShot} and directly applied to the other 12 categories without any transfer learning.
	\item \textbf{FT+TL} 
	This setting is to apply transfer learning on other categories on the best fine-tuned checkpoint on the \texttt{memc} category.
	With the help of transfer learning, we expect the PLMs have satisfying performance across different vulnerability categories.
	\item \textbf{FT+TL+SS} 
	This setting is built upon the \textbf{FT+TL} setting by allowing an additional application of \texttt{StructShot} on the model that experiences both fine-tuning and transfer learning.
	We expect a better performance of this setting over the \textbf{FT+TL} setting because of \texttt{StructShot}.
\end{itemizebullet} 

\subsection{Experimental Results: Fine-Tuning on the \texttt{memc} Category}\label{sec-ft-exp}

\subsubsection{Summary of Main Results}
The fine-tuning performance is detailed in Table \ref{tab:ft}.
We could observe that the absolute difference of F1 score for both \texttt{SN} and \texttt{SV} are both within 1\% absolute difference of the \texttt{VIEM} system.
Importantly, the amount of data we use is only 10\% of the \texttt{VIEM} system.

\begin{table}[h]
    \centering
	\caption{Fine-tuning results for the \texttt{memc} category}
	\begin{tabular}{lrrrrrr}
		\toprule
		{} &  SN precision &  SN recall &  SN f1-score &  SV precision &  SV recall &  SV f1-score \\
		\midrule
		\texttt{VIEM} (full data) &        0.9773 &     0.9916 &       0.9844 &        0.9880 &     0.9923 &       0.9901 \\
		BERT (10\% data)     &        0.9582 &     0.9808 &       0.9694 &        0.9818 &     0.9848 &       0.9833 \\
		RoBERTa (10\% data)  &        0.9567 &     0.9808 &       0.9686 &        0.9841 &     0.9844 &       0.9843 \\
		Electra (10\% data) &        0.9677 &     0.9754 &       0.9715 &        0.9830 &     0.9890 &       0.9860 \\
		\bottomrule
	\end{tabular}

	\label{tab:ft}
\end{table}

\begin{figure}[h]
    \centering
	\includegraphics[width=\linewidth]{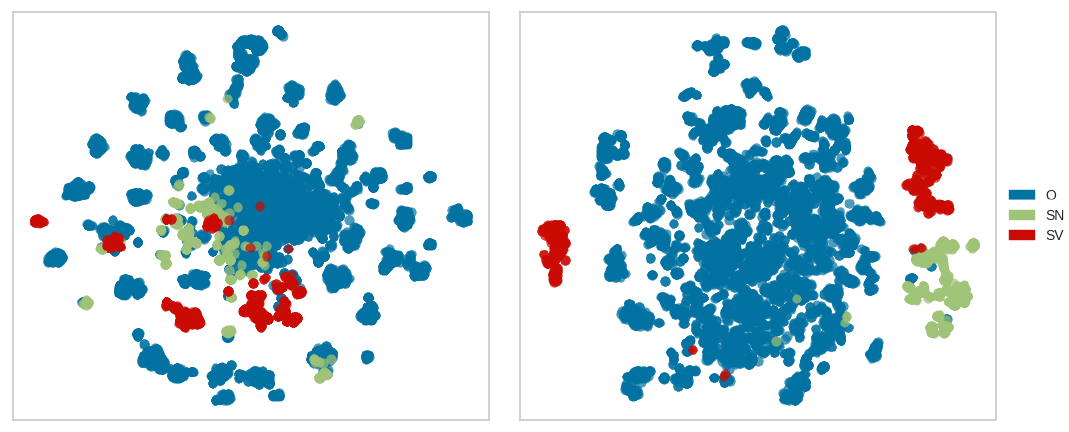}
	\caption{The t-SNE visualization of \texttt{memc} training set in token level before (left) and after (right) fine-tuning. Fine-tuning gives more compact clusters for both \texttt{SN} and \texttt{SV}, illustrating the critical role of fine-tuning.}
	\label{fig:vis}
\end{figure}

To investigate how fine-tuning helps PLMs adapt to new domain and pick up task-awareness,
we show a 2D projection of token-level hidden representation obtained on Electra for \texttt{memc} training set before and after fine-tuning in Figure \ref{fig:vis}.
We could see that much more compact clusterings of \texttt{SN} and \texttt{SV} are obtained; fine-tuning informs the PLMs with the underlying patterns of task dataset.

\subsubsection{Fine-Tuning with Different Labeled Training Sample Size}
We vary the number of labeled samples per category from 1\% to 10\% as used in \texttt{VIEM} system to see how different sizes of training set affect the performance of fine-tuning.

\begin{figure}
    \centering
    \begin{subfigure}{0.32\textwidth}
        \includegraphics[width=\linewidth]{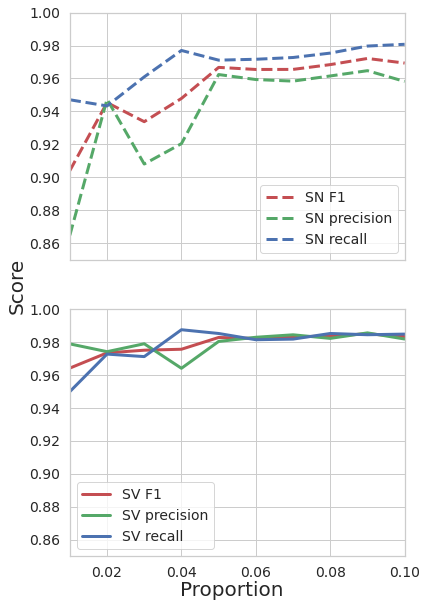}
        \caption{BERT}
        \label{fig:ft-size-bert}
    \end{subfigure}
    \begin{subfigure}{0.32\textwidth}
        \includegraphics[width=\linewidth]{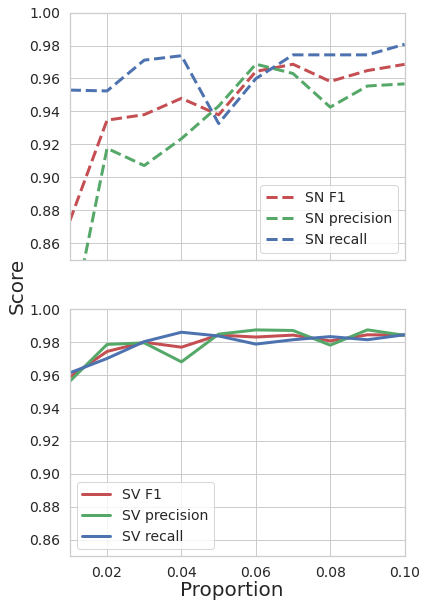}
        \caption{RoBERTa}
        \label{fig:ft-size-roberta}
    \end{subfigure}
    \begin{subfigure}{0.32\textwidth}
        \includegraphics[width=\linewidth]{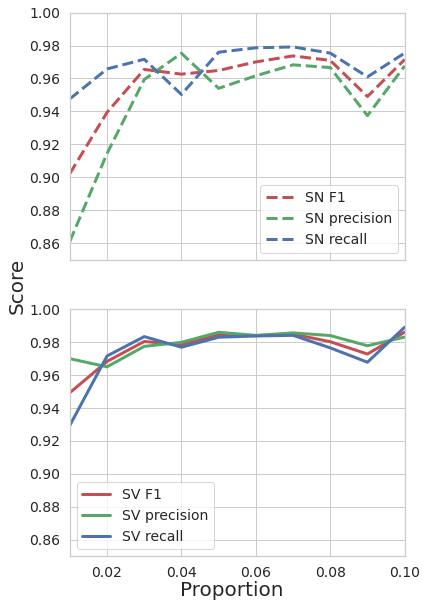} 
        \caption{Electra}
        \label{fig:ft-size-electra}
    \end{subfigure}
    \caption{The influence of training set size on fine-tuning.}
    \label{fig:ft-size}
\end{figure}

As is shown in Figure \ref{fig:ft-size}, when the proportion of labeled training samples increase from 1\% to 10\% for fine-tuning, the performance generally improves; but the improvement does not come monotonically.
For example, in cases like fine-tuning RoBERTa with 5\% of the data, the performance is worse than that of only 3\%.

There is also a disparity on the fine-tuning performance of \texttt{SN} and \texttt{SV}.
Specifically, the \texttt{SV} could have an F1 of around 95\% with as small as 1\% of the training samples available across three different models.
This does not hold for the \texttt{SN}: under the 1\% training data availability, the F1 for \texttt{SN} does not exceed 92\% for BERT; for the RoBERTa and Electra, this metric is less than 88\%.
The disparity on the performance shows that correctly tagging \texttt{SV} is easier than tagging \texttt{SN}.
This might arise from the fact that the there are naming conventions of software versions for the NER tagger to exploit while similar conventions do not exist for software names.

Another general tendency is the relation between precision and recall: across different models and the tag of interest, the recall is generally higher than precision.
This is desirable during the deployment for security applications as incorrectly tagging regular software as vulnerabilities is acceptable, only leading to more manual efforts to double check individual software, while missing vulnerabilities could leave the critical systems unattended.

\begin{figure}
    \centering
    \begin{subfigure}{0.32\textwidth}
        \includegraphics[width=\linewidth]{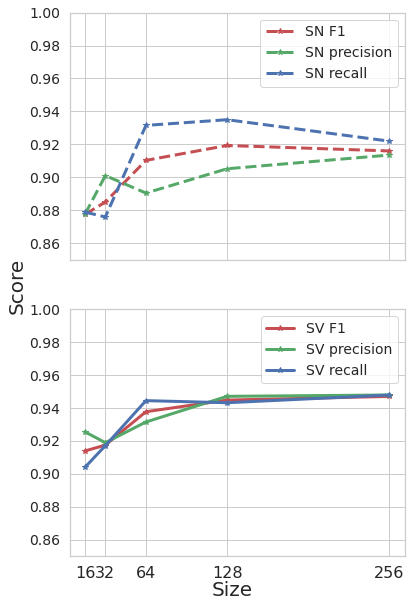}
        \caption{BERT}
        \label{fig:tl-size-bert}
    \end{subfigure}
    \begin{subfigure}{0.32\textwidth}
        \includegraphics[width=\linewidth]{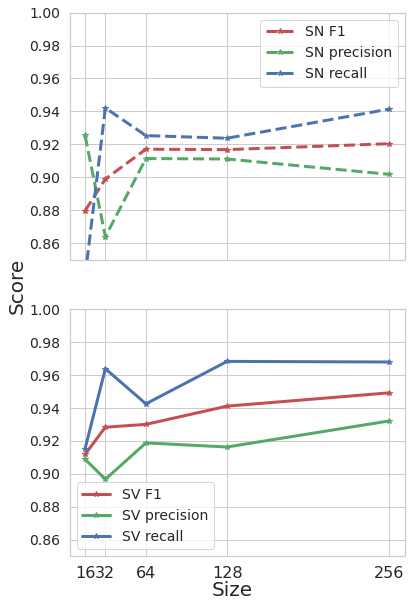}
        \caption{RoBERTa}
        \label{fig:tl-size-roberta}
    \end{subfigure}   
    \begin{subfigure}{0.32\textwidth}
        \includegraphics[width=\linewidth]{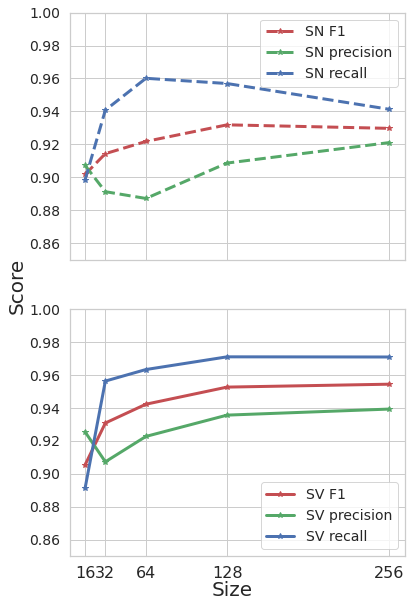} 
        \caption{Electra}
        \label{fig:tl-size-electra}
    \end{subfigure}

    \caption{The influence of training set size on transfer learning.}
    \label{fig:tl-size}
\end{figure}

\subsection{Experimental Results: Transfer Learning on the Other 12 Categories}
\label{sec:exp_transfer}

\subsubsection{Summary of the Main Results}
As we could observe from the Table \ref{tab:tl-agg}, transfer learning with the aggregation of 64 samples per category matches the performance of \texttt{VIEM} system; the absolute difference is less than 1\% in F1 score.
Despite less than 1\% of improvement with respect to our system, the \texttt{VIEM} system is trained with the entire dataset with an aggregation of 7016 samples in total (see Figure \ref{tab:stat-detail} for details; note the \texttt{VIEM} dataset merges train and valid split as its train split).
Comparing the data utilization of our system to \texttt{VIEM} system, we reduce the number of annotations by 88.80\%.

\texttt{StructShot} could help with the \emph{precision} of \texttt{SN} but struggle with the \emph{recall}.
Introducing \texttt{StuctShot} has caused a degradation in F1 score, which casts the doubt on the applicability of this system to our application.
Specifically, the absolute gap in F1 score could be up to 20\% (compared to \texttt{SN} and \texttt{SV}'s F1 scores in \textbf{FT+SS} and \textbf{FT+TL} for Electra).
This gap still exists even with additional transfer learning (\textbf{FT+TL+SS}).
Specifically, 
\begin{itemizebullet}
\item 
Comparing BERT and RoBERTa's \textbf{FT+TL+SS} with \textbf{FT+TL} in \texttt{SN} and \texttt{SV}'s precision,
adding \textbf{SS} on top of either \textbf{FT} or \textbf{FT+TL} could improve the precision for both \texttt{SN} and \texttt{SV}.
However, this is not always the case: adding \textbf{SS} to Electra's \textbf{FT+TL} actually degrades both the precision and recall for \texttt{SN} and \texttt{SV}.
\item In cases the precision does improve, as when comparing BERT and RoBERTa's \textbf{FT+TL+\allowbreak SS} with \textbf{FT+TL}, a minor improvement of precision is often at the cost of a larger degradation of recall, collectively leading to a drop in F1 score; this is especially the case for \texttt{SN}. 
\item Comparing \textbf{FT+SS} and \textbf{FT+SS+TL} for all three models' precision and recall, additional transfer learning could remedy the degradation caused by the drop of recall discussed above, but this tendency is not reverted.
\end{itemizebullet}

\subsubsection{Transfer Learning with Different Training Sample Size}
We conduct a similar performance analysis when varying the training sample size as in fine-tuning; we visualize the average performance metrics since we apply transfer learning on fine-tuned model on the other 12 categories.
As is shown in Figure \ref{fig:tl-size}, 
when the number of samples range from 32, 64 through 256, more samples do contribute to improvements on transfer learning, but the gains are less clear when we increase the training set size beyond 64 per category: the performance reaches plateau with more than 64 samples (see the F1 score of both \texttt{SN} and \texttt{SV} across three models).

We could observe a similar performance disparity of \texttt{SN} and \texttt{SV} as in fine-tuning on the \texttt{memc} category (see section \ref{sec-ft-exp}). 
Specifically, we could use as few as 16 samples per category to achieve an F1 score around 91\% while the same performance is not attainable for \texttt{SN} until we enlarge the training set to 64 (for BERT and RoBERTa) or even 128 (for Electra) samples per category.

The relation between precision and recall are consistent in transfer learning and fine-tuning: the recalls are both higher than the precision scores in most of the cases.

\subsubsection{Training Set Sampling for Transfer Learning with Single Category and 12 Category Aggregate}
The performance metrics detailed in Table \ref{tab:tl-agg} show that we could reach the performance of competitive systems by aggregating the 64 samples from each of the 12 categories, resulting in a training set In addition, we also evaluate the case where we only use the 64 examples in each category for training (and testing) without aggregation. We could see from the Table \ref{tab:tl-single} that aggregating the training samples is favorable for transfer learning, improving the system performance by 2\% to 3\%.

\begin{table*}
    \centering
	\caption{Transfer learning results averaged through 12 categories.}
	\begin{tabular}{lrrrrrr}
		\toprule
		{} &  SN precision &  SN recall &  SN f1-score &  SV precision &  SV recall &  SV f1-score \\
		\midrule
		\texttt{VIEM} (w/o transfer) &        0.8278 &     0.8489 &       0.8382 &        0.8428 &     0.9407 &       0.8891 \\
		\texttt{VIEM} (w/ transfer)  &        0.9184 &     0.9286 &       0.9235 &        0.9382 &     0.9410 &       0.9396 \\
		\multicolumn{7}{c}{BERT}\\
		FT    &        0.8759 &     0.7950 &       0.8318 &        0.8813 &     0.9341 &       0.9060 \\
        FT+SS &        0.9035 &     0.6843 &       0.7623 &        0.9001 &     0.7880 &       0.8183 \\

		FT+TL                        &        0.8945 &     0.9302 &       0.9116 &        0.9373 &     0.9354 &       0.9355 \\
        FT+TL+SS                     &        0.9060 &     0.7637 &       0.7766 &        0.9374 &     0.9135 &       0.9235 \\

		\multicolumn{7}{c}{RoBERTa}\\
		FT          				 &        0.8558 &     0.8510 &       0.8524 &        0.8822 &     0.9308 &       0.9052 \\
		FT+SS       				 &        0.8905 &     0.6810 &       0.7532 &        0.9106 &     0.8489 &       0.8735 \\
		FT+TL                        &        0.8749 &     0.9409 &       0.9063 &        0.9162 &     0.9561 &       0.9354 \\
		FT+TL+SS                     &        0.8841 &     0.7853 &       0.7989 &        0.9186 &     0.9364 &       0.9265 \\
		\multicolumn{7}{c}{Electra}\\
		FT          				 &        0.8656 &     0.8236 &       0.8416 &        0.8852 &     0.9246 &       0.9037 \\
		FT+SS       				 &        0.8692 &     0.6806 &       0.7274 &        0.8355 &     0.7428 &       0.7420 \\
		FT+TL                        &        0.9003 &     0.9443 &       0.9214 &        0.9264 &     0.9494 &       0.9369 \\
		FT+TL+SS                     &        0.8933 &     0.8263 &       0.8315 &        0.9035 &     0.9477 &       0.9233 \\
		\bottomrule
	\end{tabular}
	\label{tab:tl-agg}
\end{table*}

\begin{table*}
	\caption{Comparison of transfer learning results with training set sampled from individual and aggregate of 12 categories. The results are averaged through 12 categories.}
	\begin{tabular}{lrrrrrr}
		\toprule
		{} &  SN precision &  SN recall &  SN f1-score &  SV precision &  SV recall &  SV f1-score \\
		\midrule
		\multicolumn{7}{c}{RoBERTa}\\
		FT+TL (individual)      &        0.8483 &     0.8953 &       0.8688 &        0.8816 &     0.9342 &       0.9059 \\
        FT+TL (aggregate)       &        0.8749 &     0.9409 &       0.9063 &        0.9162 &     0.9561 &       0.9354 \\
		\multicolumn{7}{c}{Electra}\\
		FT+TL (individual)       &        0.8670 &     0.9247 &       0.8940 &        0.9015 &     0.9346 &       0.9169 \\
		FT+TL (aggregate)        &        0.9003 &     0.9443 &       0.9214 &        0.9264 &     0.9494 &       0.9369 \\
		\bottomrule
	\end{tabular}
	\label{tab:tl-single}
\end{table*}


\subsubsection{Analysis on the Two Data-Specific Challenges}
We discussed in section \ref{sec:dataset} that the \texttt{VIEM} dataset contains the challenges in \emph{context dependency} and \emph{dataset imbalance}.
Even though we do not explicitly address either of them, our experiments show that they have not largely affected the performance. 
More specifically,

\begin{itemizebullet}
    \item \textbf{Dataset Imbalance} The average proportion of sentences with entities (\texttt{SN} and \texttt{SV}) in training set is only 24.35\%; this might be a concern of performance due to the sparsity of informative tokens.
    However, as is shown in Table \ref{tab:tl-agg}, the performance of models reaches or outperforms the \texttt{VIEM} system with an average F1 score of more than 0.9, which shows that the dataset imbalance problem does not hurt the effectiveness of our system. 
    
    \item \textbf{Context Dependency} Our experiments show that transfer learning can help address the context dependency problem.
    Specifically, for the examples below, the model without fine-tuning misclassifies the tokens \textit{Network Data Loss Prevention}, \textit{Cisco Wireless LAN Controller}, and \textit{PineApp Mail-SeCure} as non-entities (\texttt{O}).
    \vspace{1em}
    \begin{itemize}
        \item[-] CVE-2014-8523 (from \texttt{csrf} category): \textit{Cross-site request forgery  (\mybox{CSRF}) vulnerability in McAfee \mybox{Network Data Loss Prevention} (NDLP) before 9.3 allows remote  \ldots}
        \item[-] CVE-2015-0690 (from \texttt{xss} category): \textit{Cross-site scripting (XSS) vulnerability \ldots on \mybox{Cisco Wireless LAN Controller} (WLC) devices \ldots aka Bug ID CSCun95178.}
        \item[-] CVE-2013-4987 (from \texttt{gainpre} category): \textit{\mybox{PineApp Mail-SeCure} before 3.70 allows remote authenticated \ldots} 
    \end{itemize}
\end{itemizebullet}

\begin{figure}[h]
    \centering
	\includegraphics[width=\linewidth]{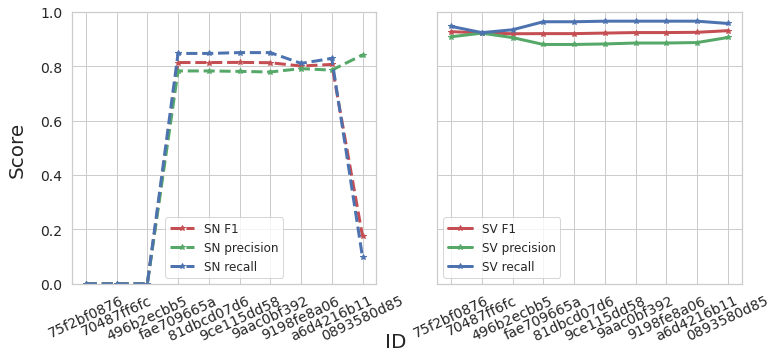}
	\caption{\texttt{StructShot} is sensitive to adversarial samples. When testing on \texttt{csrf} category, an additional sample \texttt{0893580d85} in the support set could lead to a degradation of 70\% of F1 score for \texttt{SN}.}
	\label{fig:adv}
\end{figure}

\subsubsection{Troubleshooting StructShot}
The unsatisfactory performance of \texttt{StructShot} shown in Table \ref{tab:tl-agg} alerts caution to which scenario it could be applied. 
We therefore conduct an error analysis of \texttt{StructShot} to pinpoint its weakness.
In our controlled experiment, we find that the \texttt{StructShot} is sensitive to adversarial examples.
More specifically, as is shown in Figure \ref{fig:adv}, when we manually create the support set by sequentially adding sample one after another, one single sample \texttt{0893580d85} could bring down overall performance by over 70\%.

One benefit of the \texttt{StructShot} reported in \cite{Yang2020SimpleAE} is its better capability of assigning non-entity tag (i.e. \texttt{O} tag) as these tokens do not carry unified semantic meaning.
However, because of the \emph{context dependency} illustrated in Figure \ref{fig:example}, in the NER of public security reports, non-entity tokens in one context might be entities in another one, causing confusion for the nearest neighbor-based tagger. 
We therefore suspect these confusing non-entities are the causes of the observed phenomenon.
We monitor the following error cases to validate this hypothesis. 
Specifically,
for CVE-2017-6038, the \textit{Beldn Hirschmann GECKO Lite Managed switch} is originally tagged correctly as \texttt{SN}.
However, when an additional \texttt{0893580d85} sample is introduced, all of its predictions are incorrectly reverted to \texttt{O}.
Similarly for CVE-2014-9665, the correct predictions of software names \emph{IB6131} and \emph{Infiniband Switch} are changed to \texttt{O} due to the adversarial sample \texttt{0893580d85}.
\begin{itemizebullet}
    \item CVE-2017-6038: \textit{A Cross-Site Request  Forgery issue was discovered in \mybox{Belden} \mybox{Hirschmann GECKO Lite Managed switch},  Version 2.0.00 and prior  versions .}
    \item CVE-2014-9565: \textit{Cross-site request forgery (CSRF) vulnerability in IBM Flex System EN6131 40Gb Ethernet and \mybox{IB6131} 40Gb \mybox{Infiniband Switch} 3.4.0000 and earlier.}
\end{itemizebullet}




	\section{Related Work}
\label{sec:relwork}

\subsection{Information Extraction in Public Vulnerability Database}
The \texttt{VIEM} system designed by Dong et al. is among the first batch of works that extracts information from public vulnerability database through combination of named entity recognition (NER) and relation extraction (RE) approaches.
The extracted information is used to identify vulnerabilities and thereby inconsistencies between two major security databases - CVE and NVD ~\cite{Dong2019TowardsTD}.

Prior works of the information extraction in security domain mostly focus on other aspects like dataset curation, data analysis, and visualization.
Specifically, 
Fan et al. create a C/C++ vulnerability dataset under the aid of information provided by the CVE database.
Works by Ponta et al. and Wu et al. lead to similar datasets ~\cite{Ponta2019,Fan2020}.
All of them are facilitated by the CVE database during curation while the main differences are programming language and the dataset scale.
Yamamoto et al. try to estimate the severity of vulnerability from the descriptions of CVE reports ~\cite{Yamamoto2015}.
They also propose to improve the estimation accuracy by incorporating the relative importance of reports across different years.
Pham and Dang implement an interactive tool named \texttt{CVExplorer} that enables the network and temporal view of different vulnerabilities in the CVE database ~\cite{Pham2018}.

This work follows the setting of \texttt{VIEM} system but tries to improve the vulnerability identification by minimizing the data labeling cost while matching the performance of existing systems.

\subsection{Named Entity Recognition for Computer Security}
The role of NER is not limited to serving as the first step of vulnerable software identification as in the \texttt{VIEM} system \cite{Dong2019TowardsTD}.
Indeed, there has been extensive applications of the NER in miscellaneous computer security tasks.

Feng et al. emphasize the importance of NER applications targeted at software domain for developing security applications like detecting compromised and vulnerable devices 
~\cite{Feng2018AcquisitionalRE}.
Ye et al. investigate the challenge of NER involving software, and they note that the common word polysemy is one important factor that complicates the task ~\cite{Ye2016LearningTE}.
Pandita et al. identifies software names in the description of mobile applications with NER together with other techniques; 
then they locate the sentences pertaining to security permissions with extracted information ~\cite{Pandita2013WHYPERTA}.
Sun et al. notice the sizable difference between the source of CVE records named ExploitDB and CVE database in terms of both content completeness and update time.
They, therefore, propose to automatically extract the relevant information from ExploitDB descriptions to create CVE records with NER and other technologies ~\cite{sun2021generating}.
Ma et al. cast the API extraction from software documentation as an NER task, and they leverage an LSTM-based architecture to obtain high-fidelity APIs across different libraries and languages through fine-tuning and transfer learning ~\cite{Ma2019EasytoDeployAE}.
Another work by Ye et al. attempts to use the NER to enable entity-centric applications for software engineering ~\cite{Ye2016SoftwareSpecificNE}.  However, when they reduce their training data to 10\% of the original dataset, their NER model’s performance drop from an F1 score of 78\% to 60\%.
This drop in performance highlights the difficulty of building NER systems in the few-shot setting.

Our work strives to identify vulnerabilities from public security reports, and therefore treats NER as a vehicle;
furthermore, the architecture we leverage also benefits from fine-tuning and transfer learning. 
However, we face a unique challenge of resource constraints: our system should reach competitive systems' performance with as few labeled samples as possible.  
\subsection{Few-Sample Named Entity Recognition}\label{sec:few-ner}
The study of named entity recognition dates back to Message Understanding Conference-6 (MUC-6) held in 1995, where researchers first identified the problem of recognizing named entities such as names, organizations, locations using rule-based or probabilistic approaches \cite{Grishman1996MessageUC}.
But the adoption of neural approaches and the notion of few-sample NER only happened in recent years.

Lample et al. first propose to utilize neural networks in named entity recognition task and outperforms traditional CRF-based sequence labeling models \cite{Lample2016}.
Driven by their success, neural network has become the de facto choice for named entity recognition task since then.
However, at the same time, the tension between cost of obtaining named entity annotations and data hunger nature of neural model training has gained attention.
For example, Buys et al. notice that a graphical model like HMM could outperform its neural counterparts by more than 10\% of accuracy in sequence tagging task given the access to same amount of training data \cite{Buys2018BridgingHA}.
Smaller number of parameters in HMM makes the training easier to converge than a RNN.

In order to tackle this challenge, the notion of few-shot learning, originally proposed by computer vision community \cite{Snell2017,Vinyals2016}, gets noticed by NLP practitioners.
These methods focus on inferring incoming samples' class membership based on prototypes obtained through metric learning.
Their success in image classification propels the same methods to be adapted and tested in the NER task.
Fritzler et al. are among the first to apply prototypical networks to NER task. 
However, their model treats each token independently by inferring one entity type at a time, disregarding the context of sentence \cite{Fritzler2019FewshotCI}.
Hou et al. extend this work and propose to merge specific named entity types into abstract classes (referred to as "collapsed transitions") to enable domain transfer \cite{Hou2020FewshotST}.
Yang and Katiyar borrow the idea of collapsed transitions but simplify the overall architecture \cite{Yang2020SimpleAE}. 
Rather than using dedicated architecture for few-shot learning, they first try to measure distance between tokens based on tokens' contextualized embeddings returned by PLM and then infer token's entity type based solely on nearest neighbor's types.
After this step, an optional CRF decoding module is used to account for context of tokens.
Their simple model architecture gain state-of-the-art performance compared with previous attempts.
More importantly, besides the significant boost in performance metrics, their work shows the potential of pretrained language models (PLM) in reducing annotations by leveraging universal representations of language.

Another line of works consider the alternatives to the prototype-based approaches.
Huang et al. show that self-training and active learning could help reduce data annotations \cite{Huang2020FewShotNE}.
Fries et al. provide empirical evidence of using data programming paradigm to convert rules given by the experts to named entity tags \cite{Fries2017}.
Li et al. propose a novel training protocol to make model-agnostic meta learning (MAML) framework, previously only available for sentence-level classification \cite{Bao2019}, applicable to sequence labeling (token-level classification) problem \cite{Li2020,Li2020a}.

Our work is consistent with previous works in the goal of reducing the number of training labels required.
However, different from \cite{Yang2020SimpleAE}, we discard the tag assignment scheme based on prototypes and propose the direct application of PLMs through fine-tuning and transfer learning.
The empirical evidence shows that this design choice attains better and stable performance metrics.

A distinct application scenario of few-sample learning is to test the system by inferring classes not seen in training time.
We note that, despite both striving to generalize well to unseen samples during testing,
our setting of reducing the required training examples is orthogonal to this goal following the taxonomy provided in ~\cite{Zhang2020}.
	\section{Conclusions and Future Work}\label{sec:conclusion}
In this work, we take a first attempt to investigate the problem of few-sample named entity recognition (NER) for public security vulnerability reports. 
By leveraging fine-tuning of three pre-trained language models (BERT, RoBERTa, Electra), we find that it is possible to match the performance of prior art with much less labeled training data: first, for vulnerability reports in the \texttt{memc} category, fine-tuning on PLM can match the score of prior art by using only 10\% of their training examples (or 576 labelled sentences); second, for vulnerability reports in the other 12 categories (e.g., \texttt{bypass}), we find that through fine-tuning and transfer learning, it is possible to match the scores of \cite{Dong2019TowardsTD} by using only 11.2\% of their training examples (or an aggregation of 64 labels from each category). 
As a result, few-sample learning has effectively reduced the training data required for NER for security vulnerability reports. 

It is important to notice that our work is just a beginning and there exists great potential for further improvement. 
We identify future work in the following directions. 
First, leveraging unlabelled data. 
This work only considers labelled samples of named entity tags. 
It is worth exploring how to leverage unlabelled data to further improve the performance. 
It was shown in \cite{mukherjee2020uncertainty-aware} that few-shot learning can achieve competitive performance on the GLUE dataset by leveraging models trained on a small amount of labelled data and predictions on a large amount of unlabelled examples as augmented noisy labels. 
As language models are usually pre-trained in the general domains, it can be expected that learning from unlabelled data in the security domain can help the model more quickly adapt to the in-domain task. 
Second, leveraging external knowledge base. 
Our experiment shows that it is more difficult to achieve a good F1 score for software name than software version. 
This result also meets our expectation, because there is a higher chance for a software name than software version in the testing dataset to be completely unseen. 
However, the language model may fail to bridge such gap since it is pre-trained on the general domain. 
To make the language model quickly recognize unseen software names, one approach is to leverage an external knowledge base on top of the few-sample learning model. 
It is an open question how to leverage the knowledge base to help with the few-sample learning without introducing many mismatched predictions. 
Third, by empirically observing the failure cases in transfer learning, we find that there exists some adversarial cases in the 12 sub-datasets (Section~\ref{sec:exp_transfer}), which results in a dramatic drop in the performance of few-sample learning. 
One direction of future work is thus to investigate adversarial learning for the few-shot transfer learning to improve its robustness.

	\bibliographystyle{template/splncs04.bst}
	\bibliography{citation.bib}

\begin{thebibliography}{10}
\providecommand{\url}[1]{\texttt{#1}}
\providecommand{\urlprefix}{URL }
\providecommand{\doi}[1]{https://doi.org/#1}

\bibitem{Bao2019}
Bao, Y., Wu, M., Chang, S., Barzilay, R.: Few-shot text classification with
  distributional signatures  (Aug 2019)

\bibitem{bridges2013automatic}
Bridges, R.A., Jones, C.L., Iannacone, M.D., Testa, K.M., Goodall, J.R.:
  Automatic labeling for entity extraction in cyber security. arXiv preprint
  arXiv:1308.4941  (2013)

\bibitem{brown2020language}
Brown, T.B., Mann, B., Ryder, N., Subbiah, M., Kaplan, J., Dhariwal, P.,
  Neelakantan, A., Shyam, P., Sastry, G., Askell, A., et~al.: Language models
  are few-shot learners. arXiv preprint arXiv:2005.14165  (2020)

\bibitem{Buys2018BridgingHA}
Buys, J., Bisk, Y., Choi, Y.: Bridging hmms and rnns through architectural
  transformations (2018)

\bibitem{chang2018pyramid}
Chang, J.R., Chen, Y.S.: Pyramid stereo matching network. In: Proceedings of
  the IEEE Conference on Computer Vision and Pattern Recognition. pp.
  5410--5418 (2018)

\bibitem{chen2021non}
Chen, H., Xia, M., Chen, D.: Non-parametric few-shot learning for word sense
  disambiguation. arXiv preprint arXiv:2104.12677  (2021)

\bibitem{Dodge2020FineTuningPL}
Dodge, J., Ilharco, G., Schwartz, R., Farhadi, A., Hajishirzi, H., Smith, N.A.:
  Fine-tuning pretrained language models: Weight initializations, data orders,
  and early stopping. ArXiv  \textbf{abs/2002.06305} (2020)

\bibitem{Dong2019TowardsTD}
Dong, Y., Guo, W., Chen, Y., Xing, X., Zhang, Y., Wang, G.: Towards the
  detection of inconsistencies in public security vulnerability reports. In:
  USENIX Security Symposium (2019)

\bibitem{Fan2020}
Fan, J., Li, Y., Wang, S., Nguyen, T.N.: A {C}/{C}++ {Code} {Vulnerability}
  {Dataset} with {Code} {Changes} and {CVE} {Summaries}. In: Proceedings of the
  17th {International} {Conference} on {Mining} {Software} {Repositories}. pp.
  508--512. ACM, Seoul Republic of Korea (Jun 2020).
  \doi{10.1145/3379597.3387501},
  \url{https://dl.acm.org/doi/10.1145/3379597.3387501}

\bibitem{farhang2020empirical}
Farhang, S., Kirdan, M.B., Laszka, A., Grossklags, J.: An empirical study of
  android security bulletins in different vendors. In: Proceedings of The Web
  Conference 2020. pp. 3063--3069 (2020)

\bibitem{Feng2018AcquisitionalRE}
Feng, X., Li, Q., Wang, H., Sun, L.: Acquisitional rule-based engine for
  discovering internet-of-thing devices. In: USENIX Security Symposium (2018)

\bibitem{finn2017model}
Finn, C., Abbeel, P., Levine, S.: Model-agnostic meta-learning for fast
  adaptation of deep networks. In: International Conference on Machine
  Learning. pp. 1126--1135. PMLR (2017)

\bibitem{Fries2017}
Fries, J., Wu, S., Ratner, A., Ré, C.: Swellshark: A generative model for
  biomedical named entity recognition without labeled data

\bibitem{Fritzler2019FewshotCI}
Fritzler, A., Logacheva, V., Kretov, M.: Few-shot classification in named
  entity recognition task. Proceedings of the 34th ACM/SIGAPP Symposium on
  Applied Computing  (2019)

\bibitem{gaglione2019equifax}
Gaglione~Jr, G.S.: The equifax data breach: An opportunity to improve consumer
  protection and cybersecurity efforts in america. Buff. L. Rev.  \textbf{67},
  ~1133 (2019)

\bibitem{gao2021system}
Gao, P., Liu, X., Choi, E., Soman, B., Mishra, C., Farris, K., Song, D.: A
  system for automated open-source threat intelligence gathering and
  management. arXiv preprint arXiv:2101.07769  (2021)

\bibitem{Gao2020}
Gao, T., Fisch, A., Chen, D.: Making pre-trained language models better
  few-shot learners  (Dec 2020)

\bibitem{gasmi2019information}
Gasmi, H., Laval, J., Bouras, A.: Information extraction of cybersecurity
  concepts: An lstm approach. Applied Sciences  \textbf{9}(19), ~3945 (2019)

\bibitem{Grishman1996MessageUC}
Grishman, R., Sundheim, B.: Message understanding conference- 6: A brief
  history. In: COLING (1996)

\bibitem{Hou2020FewshotST}
Hou, Y., Che, W., Lai, Y., Zhou, Z., Liu, Y., Liu, H., Liu, T.: Few-shot slot
  tagging with collapsed dependency transfer and label-enhanced task-adaptive
  projection network. In: ACL (2020)

\bibitem{hou2020few}
Hou, Y., Che, W., Lai, Y., Zhou, Z., Liu, Y., Liu, H., Liu, T.: Few-shot slot
  tagging with collapsed dependency transfer and label-enhanced task-adaptive
  projection network. arXiv preprint arXiv:2006.05702  (2020)

\bibitem{Huang2020FewShotNE}
Huang, J., Li, C., Subudhi, K., Jose, D., Balakrishnan, S., Chen, W., Peng, B.,
  Gao, J., Han, J.: Few-shot named entity recognition: A comprehensive study.
  ArXiv  \textbf{abs/2012.14978} (2020)

\bibitem{Johnson2019SurveyOD}
Johnson, J., Khoshgoftaar, T.: Survey on deep learning with class imbalance.
  Journal of Big Data  \textbf{6},  1--54 (2019)

\bibitem{Lample2016}
Lample, G., Ballesteros, M., Subramanian, S., Kawakami, K., Dyer, C.: Neural
  architectures for named entity recognition

\bibitem{Li2020a}
Li, J., Chiu, B., Feng, S., Wang, H.: Few-shot named entity recognition via
  meta-learning. {IEEE} Transactions on Knowledge and Data Engineering pp.~1--1
  (2020). \doi{10.1109/tkde.2020.3038670}

\bibitem{Li2020}
Li, J., Shang, S., Shao, L.: {MetaNER}: Named entity recognition with
  meta-learning. In: Proceedings of The Web Conference 2020. {ACM} (apr 2020).
  \doi{10.1145/3366423.3380127}

\bibitem{Ma2019EasytoDeployAE}
Ma, S., Xing, Z., Chen, C., Qu, L., Li, G.: Easy-to-deploy api extraction by
  multi-level feature embedding and transfer learning. IEEE Transactions on
  Software Engineering pp.~1--1 (2019)

\bibitem{Mosbach2020}
Mosbach, M., Andriushchenko, M., Klakow, D.: On the stability of fine-tuning
  bert: Misconceptions, explanations, and strong baselines  (Jun 2020)

\bibitem{mukherjee2020uncertainty-aware}
Mukherjee, S.S., Awadallah, A.H.: Uncertainty-aware self-training for few-shot
  text classification. In: NeurIPS 2020 (Spotlight). ACM (December 2020)

\bibitem{mulwad2011extracting}
Mulwad, V., Li, W., Joshi, A., Finin, T., Viswanathan, K.: Extracting
  information about security vulnerabilities from web text. In: 2011
  IEEE/WIC/ACM International Conferences on Web Intelligence and Intelligent
  Agent Technology. vol.~3, pp. 257--260. IEEE (2011)

\bibitem{Palmer2020}
Palmer, C.: The boeing 737 max saga: Automating failure. Engineering
  \textbf{6}(1), ~2--3 (feb 2020). \doi{10.1016/j.eng.2019.11.002}

\bibitem{Pandita2013WHYPERTA}
Pandita, R., Xiao, X., Yang, W., Enck, W., Xie, T.: Whyper: Towards automating
  risk assessment of mobile applications. In: USENIX Security Symposium (2013)

\bibitem{Pham2018}
Pham, V., Dang, T.: {CVExplorer}: {Multidimensional} {Visualization} for
  {Common} {Vulnerabilities} and {Exposures}. In: 2018 {IEEE} {International}
  {Conference} on {Big} {Data} ({Big} {Data}). pp. 1296--1301 (Dec 2018).
  \doi{10.1109/BigData.2018.8622092}

\bibitem{phang2018sentence}
Phang, J., F{\'e}vry, T., Bowman, S.R.: Sentence encoders on stilts:
  Supplementary training on intermediate labeled-data tasks. arXiv preprint
  arXiv:1811.01088  (2018)

\bibitem{Ponta2019}
Ponta, S.E., Plate, H., Sabetta, A., Bezzi, M., Dangremont, C.: A
  {Manually}-{Curated} {Dataset} of {Fixes} to {Vulnerabilities} of
  {Open}-{Source} {Software}. In: 2019 {IEEE}/{ACM} 16th {International}
  {Conference} on {Mining} {Software} {Repositories} ({MSR}). pp. 383--387.
  IEEE, Montreal, QC, Canada (May 2019). \doi{10.1109/MSR.2019.00064},
  \url{https://ieeexplore.ieee.org/document/8816802/}

\bibitem{snell2017prototypical}
Snell, J., Swersky, K., Zemel, R.S.: Prototypical networks for few-shot
  learning. arXiv preprint arXiv:1703.05175  (2017)

\bibitem{Snell2017}
Snell, J., Swersky, K., Zemel, R.S.: Prototypical networks for few-shot
  learning  (Mar 2017)

\bibitem{sun2021generating}
Sun, J., Xing, Z., Guo, H., Ye, D., Li, X., Xu, X., Zhu, L.: Generating
  informative cve description from exploitdb posts by extractive summarization.
  arXiv preprint arXiv:2101.01431  (2021)

\bibitem{sang2003conll}
Tjong Kim~Sang, E.F., De~Meulder, F.: Introduction to the conll-2003 shared
  task: Language-independent named entity recognition. In: Proceedings of the
  Seventh Conference on Natural Language Learning at HLT-NAACL 2003 - Volume 4.
  p. 142–147. CONLL '03, Association for Computational Linguistics, USA
  (2003). \doi{10.3115/1119176.1119195},
  \url{https://doi.org/10.3115/1119176.1119195}

\bibitem{Vinyals2016}
Vinyals, O., Blundell, C., Lillicrap, T., Kavukcuoglu, K., Wierstra, D.:
  Matching networks for one shot learning  (Jun 2016)

\bibitem{Wang2019GeneralizingFA}
Wang, Y., Yao, Q., Kwok, J., Ni, L.: Generalizing from a few examples: A survey
  on few-shot learning. arXiv: Learning  (2019)

\bibitem{Wolf2019HuggingFacesTS}
Wolf, T., Debut, L., Sanh, V., Chaumond, J., Delangue, C., Moi, A., Cistac, P.,
  Rault, T., Louf, R., Funtowicz, M., Brew, J.: Huggingface's transformers:
  State-of-the-art natural language processing. ArXiv  \textbf{abs/1910.03771}
  (2019)

\bibitem{Yamamoto2015}
Yamamoto, Y., Miyamoto, D., Nakayama, M.: Text-{Mining} {Approach} for
  {Estimating} {Vulnerability} {Score}. In: 2015 4th {International} {Workshop}
  on {Building} {Analysis} {Datasets} and {Gathering} {Experience} {Returns}
  for {Security} ({BADGERS}). pp. 67--73. IEEE, Kyoto, Japan (Nov 2015).
  \doi{10.1109/BADGERS.2015.018},
  \url{http://ieeexplore.ieee.org/document/7809535/}

\bibitem{Yang2020SimpleAE}
Yang, Y., Katiyar, A.: Simple and effective few-shot named entity recognition
  with structured nearest neighbor learning. ArXiv  \textbf{abs/2010.02405}
  (2020)

\bibitem{Ye2016SoftwareSpecificNE}
Ye, D., Xing, Z., Foo, C.Y., Ang, Z.Q., Li, J., Kapre, N.: Software-specific
  named entity recognition in software engineering social content. 2016 IEEE
  23rd International Conference on Software Analysis, Evolution, and
  Reengineering (SANER)  \textbf{1},  90--101 (2016)

\bibitem{Ye2016LearningTE}
Ye, D., Xing, Z., Foo, C.Y., Li, J., Kapre, N.: Learning to extract api
  mentions from informal natural language discussions. 2016 IEEE International
  Conference on Software Maintenance and Evolution (ICSME) pp. 389--399 (2016)

\bibitem{Zhang2020}
Zhang, T., Wu, F., Katiyar, A., Weinberger, K.Q., Artzi, Y.: Revisiting
  few-sample bert fine-tuning  (Jun 2020)

\bibitem{Zhou2020ImprovingSB}
Zhou, C., Li, B., Sun, X.: Improving software bug-specific named entity
  recognition with deep neural network. J. Syst. Softw.  \textbf{165},  110572
  (2020)

\end{thebibliography}
	\newpage
	\appendix
\section{Dataset Statistics}
Table \ref{tab:stat-detail} is the detailed version of Table \ref{tab:stat}.
The valid split for categories other than \texttt{memc} is 10\% sample of official train set. 
"Sentence-level entity proportion" refers to the proportion of sentences that have \texttt{SN} or \texttt{SV}, and "Token-level entity proportion" is proportion of \texttt{SN} and \texttt{SV} with respect to given dataset split. 
These proportions reflect the dataset imbalance in different levels.

\begin{table*}
	\centering
	\caption{Detailed statistics of \texttt{VIEM} dataset.}
	\begin{tabular}{cccccc}
		\toprule
		Category & Split &  Number of sample &  \multicolumn{1}{p{3cm}}{\centering Sentence-level\\ entity proportion} &  \multicolumn{1}{p{3cm}}{\centering Token-level entity \\ proportion (\texttt{SN})} &  \multicolumn{1}{p{3cm}}{\centering Token-level entity \\ proportion (\texttt{SV})} \\
		\midrule
		\multirow{3}{*}{\texttt{memc}} & train &              5758 &                                                                           0.5639 &                                                                           0.0613 &                                                                           0.0819 \\
		& valid &              1159 &                                                                           0.3287 &                                                                           0.0368 &                                                                           0.0807 \\
		& test &              1001 &                                                                           0.4555 &                                                                           0.0559 &                                                                           0.0787 \\
		\cline{1-6}
		\multirow{3}{*}{\texttt{bypass}} & train &               652 &                                                                           0.2239 &                                                                           0.0314 &                                                                           0.0431 \\
		& valid &               162 &                                                                           0.2469 &                                                                           0.0367 &                                                                           0.0423 \\
		& test &               610 &                                                                           0.2902 &                                                                           0.0456 &                                                                           0.0531 \\
		\cline{1-6}
		\multirow{3}{*}{\texttt{csrf}} & train &               521 &                                                                           0.2399 &                                                                           0.0207 &                                                                           0.0347 \\
		& valid &               130 &                                                                           0.2846 &                                                                           0.0251 &                                                                           0.0397 \\
		& test &               415 &                                                                           0.3181 &                                                                           0.0321 &                                                                           0.0464 \\
		\cline{1-6}
		\multirow{3}{*}{\texttt{dirtra}} & train &               619 &                                                                           0.2359 &                                                                           0.0172 &                                                                           0.0219 \\
		& valid &               155 &                                                                           0.1871 &                                                                           0.0180 &                                                                           0.0316 \\
		& test &               646 &                                                                           0.2879 &                                                                           0.0197 &                                                                           0.0220 \\
		\cline{1-6}
		\multirow{3}{*}{\texttt{dos}} & train &               396 &                                                                           0.2273 &                                                                           0.0212 &                                                                           0.0405 \\
		& valid &                99 &                                                                           0.2020 &                                                                           0.0234 &                                                                           0.0419 \\
		& test &               484 &                                                                           0.2624 &                                                                           0.0189 &                                                                           0.0331 \\
		\cline{1-6}
		\multirow{3}{*}{\texttt{execution}} & train &               413 &                                                                           0.2639 &                                                                           0.0228 &                                                                           0.0358 \\
		& valid &               103 &                                                                           0.2718 &                                                                           0.0314 &                                                                           0.0302 \\
		& test &               639 &                                                                           0.2598 &                                                                           0.0273 &                                                                           0.0357 \\
		\cline{1-6}
		\multirow{3}{*}{\texttt{fileinc}} & train &               546 &                                                                           0.2857 &                                                                           0.0175 &                                                                           0.0185 \\
		& valid &               137 &                                                                           0.3869 &                                                                           0.0259 &                                                                           0.0222 \\
		& test &               683 &                                                                           0.3133 &                                                                           0.0206 &                                                                           0.0215 \\
		\cline{1-6}
		\multirow{3}{*}{\texttt{gainpre}} & train &               323 &                                                                           0.2229 &                                                                           0.0243 &                                                                           0.0430 \\
		& valid &                80 &                                                                           0.3250 &                                                                           0.0357 &                                                                           0.0723 \\
		& test &               577 &                                                                           0.2114 &                                                                           0.0191 &                                                                           0.0311 \\
		\cline{1-6}
		\multirow{3}{*}{\texttt{httprs}} & train &               550 &                                                                           0.1891 &                                                                           0.0127 &                                                                           0.0217 \\
		& valid &               137 &                                                                           0.1241 &                                                                           0.0077 &                                                                           0.0124 \\
		& test &               411 &                                                                           0.2360 &                                                                           0.0175 &                                                                           0.0304 \\
		\cline{1-6}
		\multirow{3}{*}{\texttt{infor}} & train &               305 &                                                                           0.2459 &                                                                           0.0326 &                                                                           0.0354 \\
		& valid &                76 &                                                                           0.3158 &                                                                           0.0187 &                                                                           0.0282 \\
		& test &               509 &                                                                           0.2358 &                                                                           0.0227 &                                                                           0.0348 \\
		\cline{1-6}
		\multirow{3}{*}{\texttt{overflow}} & train &               396 &                                                                           0.2475 &                                                                           0.0217 &                                                                           0.0326 \\
		& valid &                98 &                                                                           0.2143 &                                                                           0.0185 &                                                                           0.0230 \\
		& test &               454 &                                                                           0.2819 &                                                                           0.0216 &                                                                           0.0343 \\
		\cline{1-6}
		\multirow{3}{*}{\texttt{sqli}} & train &               538 &                                                                           0.2565 &                                                                           0.0145 &                                                                           0.0141 \\
		& valid &               134 &                                                                           0.2836 &                                                                           0.0194 &                                                                           0.0151 \\
		& test &               685 &                                                                           0.2423 &                                                                           0.0171 &                                                                           0.0181 \\
		\cline{1-6}
		\multirow{3}{*}{\texttt{xss}} & train &               357 &                                                                           0.2829 &                                                                           0.0203 &                                                                           0.0289 \\
		& valid &                89 &                                                                           0.3708 &                                                                           0.0276 &                                                                           0.0386 \\
		& test &               562 &                                                                           0.2046 &                                                                           0.0219 &                                                                           0.0363 \\
		\bottomrule
	\end{tabular}
	\label{tab:stat-detail}
\end{table*}

\end{document}